\documentclass{article}

     \usepackage[preprint]{neurips_2020}

\usepackage[utf8]{inputenc} 
\usepackage[T1]{fontenc}    
\usepackage{hyperref}       
\usepackage{url}            
\usepackage{booktabs}       
\usepackage{amsfonts}       
\usepackage{nicefrac}       
\usepackage{microtype}      
\usepackage{mathtools}
\usepackage{mathrsfs}
\usepackage{cases}

\usepackage{color,xcolor}

\usepackage{multirow}
\usepackage{natbib}
\setcitestyle{square,numbers,comma}

\title{Noise Homogenization \\via Multi-Channel Wavelet Filtering \\for High-Fidelity Sample Generation in GANs}

\author{%
Shaoning Zeng \\ 
\textit{zsn@outlook.com}
\And 
Bob Zhang \thanks{The corresponding author who supervised this work.}  \\ 
\textit{bobzhang@um.edu.mo}  
\AND 
\\ 
Department of Computer and Information Science \\
University of Macau, Taipa, Macau, China \\
}

\begin{document}

\maketitle

\begin{abstract}
  In the generator of typical Generative Adversarial Networks (GANs), a noise is inputted to generate fake samples via a series of convolutional operations. However, current noise generation models merely relies on the information from the pixel space, which increases the difficulty to approach the target distribution. Fortunately, the long proven wavelet transformation is able to decompose multiple spectral information from the images. In this work, we propose a novel multi-channel wavelet-based filtering method for GANs, to cope with this problem. When embedding a wavelet deconvolution layer in the generator, the resultant GAN, called WaveletGAN, takes advantage of the wavelet deconvolution to learn a filtering with multiple channels, which can efficiently homogenize the generated noise via an averaging operation, so as to generate high-fidelity samples. We conducted benchmark experiments on the Fashion-MNIST, KMNIST and SVHN datasets through an open GAN benchmark tool. The results show that WaveletGAN has excellent performance in generating high-fidelity samples, thanks to the smallest FIDs obtained on these datasets. 
\end{abstract}

\section{Introduction}
In Generative Adversarial Networks (GANs) \cite{Goodfellow2014Generative}, the generator $G$ is designed to capture the data distribution as accurately as possible, through the adversarial training by the discriminator $D$, which will ultimately fail to distinguish the generated samples from the real ones. In this adversarial model, the problem getting the most attention is on how to efficiently train the model. This is because the only way to generate the fake samples with adequate fidelity used to depend on the fine tuning process after the discrimination by $D$. For this reason, a bunch of research effort has been devoted to improve the training from the two components, $G$ and $D$. We agree and argue that orchestrating the network architectures is one of the most promising solutions for this problem.  

Almost every single aspect of a GAN has been exploited to enhance the sample generation, including architectures, loss function, regularization, normalization, etc. First of all, most of the cutting-edge deep CNN models have been incorporated as the backbone network of GANs, to form $G$ and/or $D$. Among them, for example, ResNet \cite{He2016Deep} has been popular and showed a promising performance as the backbone network \cite{Miyato2018Spectral,Choi2018Stargan}. The well-known loss functions like Least Squares \cite{Mao2017Least}, Hinge \cite{Tran2018Dist}, and Wasserstein \cite{Arjovsky2017Wasserstein} are found in GANs for different purposes. Innovations on regularization \cite{Fedus2018Many,Gulrajani2017Improved} and normalization \cite{Miyato2018Spectral,Gulrajani2017Improved} for improving GANs have become common occurrences as well. Recently, many interesting strategies in representation learning, including self-supervised learning \cite{Chen2019Self} and self-attention learning \cite{Zhang2019Self}, have been incorporated to design more promising GAN architectures. 
However, all these explorations remain on one single dimension of spectral analysis, i.e., the pixels of images. Many other signal processing techniques that support powerful spectral decomposition, e.g., using wavelets to analyze frequency information \cite{Khan2018Learning}, are likely to enhance spectral decomposition for sample generation by multi-channel analysis. This is so far an open question to be answered.       

\begin{figure*}
  \begin{center}
    \includegraphics[width=0.95\textwidth]{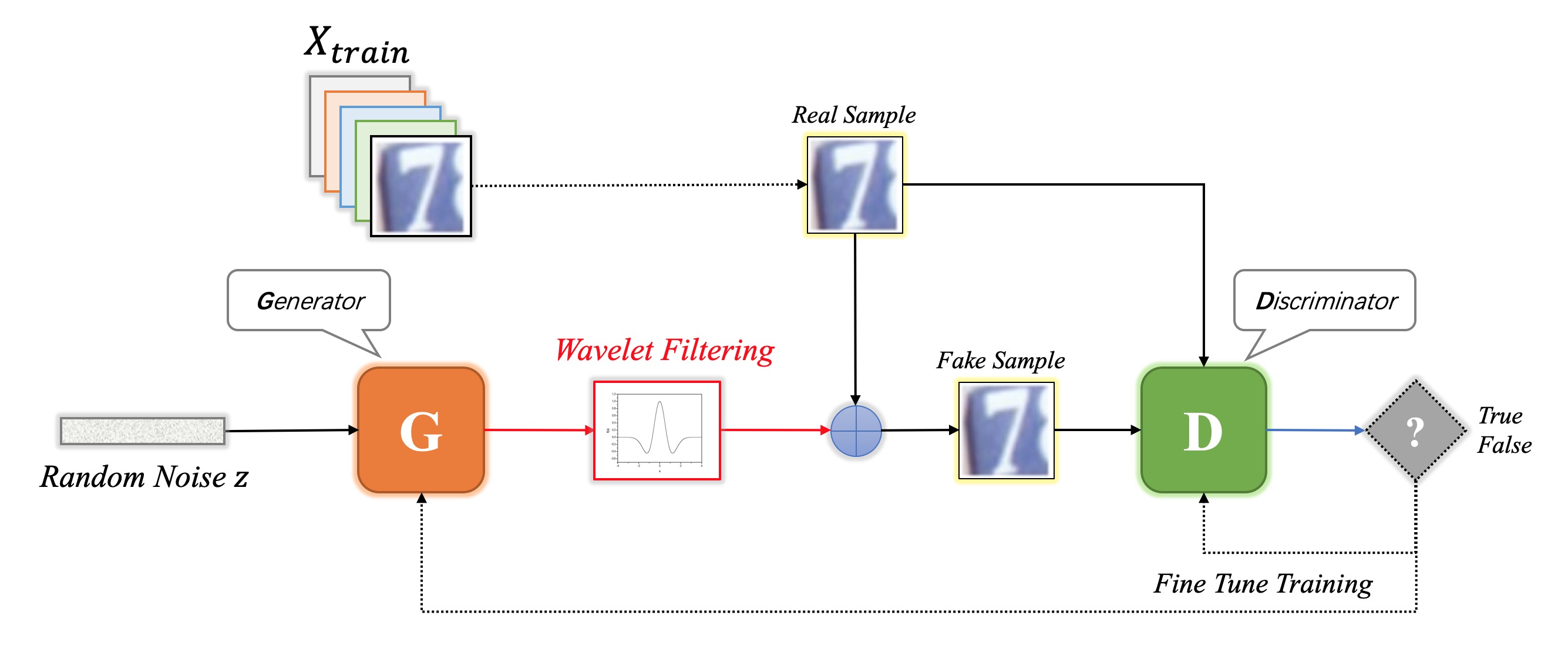}
  \end{center}
  \caption{\small WaveletGAN architecture using wavelet filtering to homogenize the generated noise.}
  \label{fig:Idea}
\end{figure*}

In this work, we firstly introduce a multi-channel wavelet-based filtering to orchestrate a novel GAN architecture, which is named WaveletGAN. The idea is easy to understand and simple to implement, as shown in Figure \ref{fig:Idea}. When the noise is generated by G, a multi-channel wavelet filtering is injected to perform spectral decomposition that considers both scale and frequency. The fake samples are then generated by incorporating this additional information. The filtering is simply implemented as a pluggable \textit{WaveletDeconv} layer, which supports most of the GAN architectures and whose details will be explained later. In this way, we intent to let the outputted noise go through a homogenization processing, so as to improve the quality of the generated samples. This is confirmed by the lowest Fréchet Inception Distances (FIDs) \cite{Heusel2017Gans} obtained by WaveletGAN in our benchmarking experiments on multiple image datasets. Our contributions in this work include: 
\begin{itemize}
  \item[$\bullet$]  Proposing a multi-channel wavelet filtering method in the generator, which excels most of the current GANs in generating high-fidelity samples. 
  \item[$\bullet$]  Giving a simple implementation, as the \textit{WaveletDeconv} layer supports almost all types of GANs, by averaging multiple spectral channels. It is intented to be a very cost-effective operation to homogenize the noise and fully utilize the enriched information, despite the fact that it is difficult to explore the most useful information.   
  \item[$\bullet$]  Demonstrating the performance of WaveletGAN through $compare\_gan$, an open-sourced GAN benchmarking tool by Google Research\footnote{Open GAN benchmarking tool - https://github.com/google/compare\_gan}, on the Fashion-MNIST \cite{Xiao2017Fashion}, KMNIST \cite{Clanuwat2018Kmnist} and Street View House Numbers (SVHN) \cite{Netzer2011Reading} datasets. The lowest FIDs are obtained by our WavletGAN, which are superior to multiple current state-of-the-art GANs. 
\end{itemize}
  
The paper is organized using the following structure. The related studies, mainly on GANs and wavelets in neural networks, are reviewed in Sec. \ref{Related}. Afterwards, the architectures of WaveletGAN, as well as its formulation and analysis, are presented in Sec. \ref{Method}. Sec. \ref{Experiments} demonstrates the benchmarking settings and the experimental results. We have the conclusions and the future work in Sec. \ref{Conclusions}.

\section{Related Work}
\label{Related}


\subsection{Generative Adversarial Networks}
Improvements to GANs fall into three categories, including carefully selecting a good loss function \cite{Arjovsky2017Wasserstein,Mao2017Least}, distinctively considering regularization \cite{Fedus2018Many} and/or normalization \cite{Miyato2018Spectral,Gulrajani2017Improved}, and painstakingly designing the neural architecture \cite{He2016Deep,Miyato2018Spectral,Radford2016DCGAN}, for both the generator and the discriminator. Furthermore, many of the current improvements cover more than one of the categories. For example, regularizations like gradient penalty \cite{Gulrajani2017Improved,Roth2017Stabilizing}, consensus constraint \cite{Mescheder2017Numerics}, 
and spectral normalization \cite{Miyato2018Spectral} are specifically applied to the discriminator, so as to mitigate those gradient vanishing issues in training GANs. Also, there are similar strategies particularly for the generator, like applying spectral normalization \cite{Miyato2018Spectral} to improve training dynamics \cite{Zhang2019Self}. This makes things complicated and extremely hard. 

As a descendant trend of deep neural networks \cite{Goodfellow2014Generative}, architecture design for GANs has gained the most attention. Some of the current methods merely focused on discriminator learning \cite{Zhang2020Consistency,Nguyen2017Dual}, while the others were more specific on the generator \cite{Hoang2018Mgan}, or both \cite{Tran2019Self}. 
In correspondence, the GAN architecture with multiple generators has also been proposed \cite{Hoang2018Mgan}. That being said, there are a lot of complicated models proposed by stacking multiple architectures \cite{Nguyen2017Dual,Hoang2018Mgan}. 
Among these areas, there is another trend specific on the signal conditions, i.e., conditional training with labels \cite{Zhang2019Self} 
and unsupervised GANs trained without prior labels \cite{Chen2019Self}. More recently, the self-supervised constrains \cite{Chen2019Self,Tran2019Self} and the self-attention mechanism \cite{Zhang2019Self} were also presented to improve GANs. 

Besides these, architectures that manipulate the noise generation also attracts a lot of attentions. The multiplicative noise was introduced to reduce the uncertainty in GANs \cite{Di2017Multiplicative}. A decoder-encoder network was particularly designed to map the random noise vectors to informative ones and feed them to the generator of the adversarial networks \cite{Zhong2020Generative}. A smooth generator was proposed by manipulating the output noise to cope with the perturbation during sample generation \cite{Guo2019Smooth}. Therefore, we believe that noise manipulation provides a promising potential to produce high-fidelity sample generation in GANs. However, current options have been staying in the pixel space of noise.  


\subsection{Wavelets in neural networks}

Wavelet transformation, known as a well-proven and fully-utilized time-frequency signal analysis tool \cite{Daubechies1990Wavelet}, has also been playing an important role in constructing neural networks \cite{Zhang1992Wavelet}. For example, it is able to approximate arbitrary nonlinear functions \cite{Zhang1992Wavelet} and replace the radial basis function \cite{Ho2001Fuzzy} in neural networks. Besides these, fuzzy wavelet neural networks \cite{Davanipoor2011Fuzzy} included a wavelet function to formulate dynamic systems for a better performance. Recently, a deep wavelet network was proposed for image classification \cite{Said2016Deep}. Multi-level wavelet deep CNNs were designed for image restoration \cite{Liu2018Multi} and fault diagnosis \cite{Han2018Multi}. Wavelet transform has been integrated into a deep RNN model for natural gas demand forecasting \cite{Su2019Hybrid}. Furthermore, various types of wavelet-like deep auto-encoders have been proposed to accelerate deep neural networks \cite{Chen2018Learning}, as well as different applications, including image classification \cite{Luo2018Wavelet}, intelligent fault diagnosis \cite{He2020Deep}, medical imaging \cite{Mallick2019Brain}, spherical signal detection \cite{Wu2019Rotated}, etc. Graph wavelets were constructed to sparsely represent a given class of signals in deep learning \cite{Rustamov2013Wavelets}. 

However, no exploration has occurred around utilizing wavelets to improve image generation in GANs. It is well-known that wavelet transformation is capable of filtering and performing deconvolution for signal processing \cite{Starck1994Filtering}, for example, capturing the addition frequency information. We believe that it is helpful to homogenize the noise production in the generator, which is likely to simulate the distribution easier and be accepted by the discriminator. Here, we design our proposed WaveletGAN based on this assumption.       

\section{WaveletGAN: Noise Homogenization for High-Fidelity Generation}
\label{Method}

\subsection{GAN architectures}

The formulation of GANs in the convolutional form \cite{Goodfellow2014Generative} is 
\begin{equation}
	\min _{G} \max _{D} V(G, D) = \mathrm{E}_{\boldsymbol{x} \sim q_{data}}[\log D(\boldsymbol{x})]+\mathrm{E}_{\boldsymbol{z} \sim p_{G}} [\log  (1-D (G(\boldsymbol{z}) ) ) ],
\end{equation}
where $q_{data}$ is the data distribution, and $p_G$ is the (model) generator distribution to be learned through the adversarial min-max optimization by the fine-tuning game between $G$ and $D$. Given a noise $\boldsymbol{z}$, $\boldsymbol{x}^{\prime} = G(\boldsymbol{z})$ is the fake sample generated by $G$, which is then to be discriminated by $D$. In a simple implementation, the noise $\boldsymbol{z}$ is the output of the generator upon an initial input noise $\boldsymbol{z}_0$, a.k.a.  $\boldsymbol{z} = G(\boldsymbol{z}_0)$. Then, the generated sample can be obtained by adding the output noise to the real sample, i.e., $\boldsymbol{x}^{\prime} = G(\boldsymbol{z}_0) + x$. In this way, the formulation of this kind of simple GANs becomes 
\begin{equation}
	\min _{G} \max _{D} V(G, D) = \mathrm{E}_{\boldsymbol{x} \sim q_{\text {data }}}[\log D(\boldsymbol{x})]+\mathrm{E}_{\boldsymbol{z} \sim p_{G}}[\log (1-D(\boldsymbol{x}+G(\boldsymbol{z_0})))].
\end{equation}
Different from the methods aiming at enhancing the discrimination of $D$, a few of the new mechanisms have been revealed to manipulate the output noise $\boldsymbol{z}$ \cite{Di2017Multiplicative,Zhong2020Generative,Guo2019Smooth}. Here, we are going to take advantage of the frequency decomposed by the wavelet transformations.   

The key step (one and only) in our proposed method is an additional wavelet deconvolution layer, which performs a multi-channel wavelet filtering operation to homogenize the generated noise. Typically, the generator network works on a random initialized noise $\boldsymbol{z_0}$, i.e., the input to the generator. In conventional GANs, the generated noise was added to the real images to generate the fake samples, before being fed to the discriminator. Technically, the wavelet deconvolution supports random forms of noise, as a signal to be analyzed by wavelets, hence our method supports any GAN models. Indeed, it can be implemented \textbf{in one line of code}\footnote{See the accompanying code for this work in https://github.com/zengsn/compare\_gan.
}. For simplicity, we give an implementation based on the GAN architecture provided in \cite{Miyato2018Spectral}, which used ResNet \cite{He2016Deep} as the backbone network, for the sake of demonstrating the architecture of WaveletGAN. 

\begin{table}
  \caption{ResNet-based GAN architectures with wavelet deconvolution.}
  \label{table:resnetGAN}
  \center \small
  \begin{tabular}{cc}
    \begin{tabular}{ll}
      \toprule
      Generator & Discriminator  \\
      \midrule
      $\boldsymbol{z}_0 \in \mathbb{R}^{128} \sim \mathcal{N}(0, I)$ 	& $\boldsymbol{x} \in \mathbb{R}^{28 \times 28}$ \\
      dense, 7 $\times$ 7 $\times$ 256 			& ResBlock down 128 \\
      ResBlock up 256 						& ResBlock down 128 \\
      ResBlock up 256 						& ReLU \\
      BN, ReLU, 3$\times$3 conv 3 				& Global sum pooling \\
      \textit{WaveletDeconv, 5, average} 			& dense $\rightarrow$ 1 \\
      Sigmod 								& \\
      & \\
      \bottomrule 
    \end{tabular}
    &
    \begin{tabular}{ll}
      \toprule
      Generator & Discriminator  \\
      \midrule
      $\boldsymbol{z}_0 \in \mathbb{R}^{128} \sim \mathcal{N}(0, I)$ & $\boldsymbol{x} \in \mathbb{R}^{32 \times 32 \times 3}$ \\
      dense, $4 \times 4 \times 256$ 				& ResBlock down 128 \\
      ResBlock up 256 						& ResBlock down 128 \\
      ResBlock up 256 						& ResBlock 128 \\
      ResBlock up 256 						& ResBlock 128 \\
      BN, ReLU, 3$\times$3 conv 3 				& ReLU \\
      \textit{WaveletDeconv, 5, average} 			& Global sum pooling \\
      Sigmod 								& dense $\rightarrow$ 1\\
      \bottomrule 
    \end{tabular} \\
    \noalign{\smallskip}
    (a). Architecture for FMNIST and KMNIST. & (b). Architecture for SVHN. \\
  \end{tabular}
\end{table}

We implement two versions of ResNet architectures, according to \cite{Miyato2018Spectral}, for two types of input data. The generators accept the same input of the initial noise $\boldsymbol{z}_0 \in \mathbb{R}^{128} \sim \mathcal{N}(0, I)$. However, the target outputs depend on the datasets. The first one, as shown in Table \ref{table:resnetGAN}-(a), is for the training data $x \in \mathbb{R}^{28 \times 28}$, while the other one supports the RGB images $\boldsymbol{x} \in \mathbb{R}^{32 \times 32 \times 3}$, as shown in Table \ref{table:resnetGAN}-(b). Technically, it is all about the backbone networks, i.e., ResNet, in the architecture, which contains 2 or 3 \textit{ResBlock}s, respectively. However, the backbone network is easily alternated to support other scales of input. We omit this part because our wavelet-based noise homogenization has nothing to do with the network, but only depends on the noise generations. As shown in Table \ref{table:resnetGAN}, the \textit{WaveletDeconv} layer is right after the output noise finalized by a $3 \times 3$ convolution. We set the number of channels to be a fixed value of $5$, which in turn produces 5 channels of analyzed results. An average operation is then performed upon these resultant signals, as an operation of noise homogenization, to compose the new output noise for the following sample generation.   


\subsection{Multi-channel wavelet filtering}
\label{waveletFiltering}

The noise homogenization is implemented by introducing multi-channel wavelet filtering, right after the outputted noise by the generator. First of all, we introduce the continuous wavelet transformation (CWT) \cite{Hammond2011Wavelets} to the signal, i.e., the noise $\boldsymbol{z}$ outputted by the generator (see Figure \ref{fig:Idea}). CWT is defined by a mother wavelet function $\mathbf{\Psi}$, which is scaled to form the wavelet functions. This transformation is usually good at processing a time series signal $x(t)$ (where $t = 1\dots T$) \cite{Khan2018Learning}, convolved with a specific wavelet function. According to the wavelet theory \cite{Daubechies1990Wavelet}, the prerequisite to select the mother wavelet function is making sure it has small local support and satisfies the zero mean and normalization properties.
We choose the Mexican Hat wavelet as the mother wavelet, since it is common for combining wavelet transformations with convolutional neural networks \cite{Wiatowski2017Mathematical,Ji2019Research,Lee2019Application}, whose definition can be derived from a Gaussian:   

\begin{equation}
	\psi(t)=\frac{2}{\pi^{1 / 4} \sqrt{3 \sigma}}\left(\frac{t^{2}}{\sigma^{2}}-1\right) e^{-\frac{t^{2}}{\sigma^{2}}}.
\end{equation}

By scaling and translating, this formulation can be rewritten to a new form of the wavelet function $\psi_{s,b}$:

\begin{equation}
	\psi_{s, b}(t)=\frac{1}{\sqrt{s}} \psi\left(\frac{t-b}{s}\right),
\end{equation} 
where $s > 0$ is the scale coefficient and $b$ is the offset to translate. Then, the CWT of a signal, i.e., decomposition of $\boldsymbol{z}$ according to $s$ and $b$, becomes a new formulation as follows: 

\begin{equation}
	W_{\boldsymbol{z}}(s, b)=\int_{-\infty}^{\infty} \frac{1}{\sqrt{s}} \psi\left(\frac{t-b}{s}\right) \boldsymbol{z}(t) dt. 
\end{equation}

In this way, the signal $\boldsymbol{z}$ will be transformed from one single dimension of domain ($t$) to two dimensions ($t$ and $s$) by convolution with a wavelet function $\mathbf{\Psi}$ at each scale $s$. This is the basis of how we can implement a multi-channel wavelet filtering layer in a deep convolutional neural network. It is noted that this kind of wavelet deconvolution layer (a.k.a. \textit{WaveletDeconv}) has been implemented for the time series-based phone recognition task \cite{Khan2018Learning}, which showed that the gradients of the scales can be learned using back-propagation. 

Denote $E$ as a differentiable loss function to be minimized by the neural network, which is typically the categorical cross entropy, the gradient, yielded by the wavelet deconvolution layer, with respect to the scale parameter $s$ for training the network is:

\begin{equation}
\label{eq:Gradient}
	\frac{\delta E}{\delta s_{i}}=\sum_{k=1}^{K} \frac{\delta E}{\delta \psi_{s_{i}, k}} \frac{\delta \psi_{s_{i}, k}}{\delta s_{i}},
\end{equation} 
where $K$ is the width of the filter that can be learned during training. At this time, the scales can be dynamically updated according to the gradients:   
\begin{equation}
	s_{i}^{\prime}=s_{i}-\gamma \frac{\delta E}{\delta s_{i}}, 
\end{equation}
where $\gamma$ is the learning rate of the optimizer. We omit the full solution of Eq. \ref{eq:Gradient}, which can be found in \cite{Khan2018Learning}. We reuse the same implementation of \textit{WaveletDeconv} from \cite{Khan2018Learning} in our model. Consequently, the new formulation for our WaveletGAN becomes:

\begin{equation}
	\min _{G} \max _{D} V(G, D) = \mathrm{E}_{\boldsymbol{x} \sim q_{\text {data }}}[\log D(\boldsymbol{x})]+\mathrm{E}_{\boldsymbol{z} \sim p_{G}}\left[\log \left(1-D\left(\boldsymbol{x}+W(G(\boldsymbol{z_0}))\right)\right)\right].
\end{equation}

However, there are two-fold differences between our \textit{WaveletDeconv} and the original one. Firstly, we implement \textit{WaveletDeconv} for image generation in GANs, which has no time series information. Secondly, we support multi-channel filtering, rather than single-channel filtering, followed by an additional averaging operation to combine multiple scales of signals into one target noise signal, and therefore homogenize the output noise. In this way, the noises before and after processing are in an identical shape, and hence our WaveletGAN has no dependency on the network architectures. WaveletGAN simply acts like a novel preprocessing technique, which is therefore very easy to implement and even can be written in one line of code, to support most of the GAN architectures. 

\section{Experiments}
\label{Experiments}

We conducted a set of experiments and ran them on GPUs using the GAN benchmarking tool $compare\_gan$\footnote{It supports both GPUs and Google TPU, but we ran the benchmarking experiments on GPUs only.}, to evaluate the performance of our WaveletGAN. As aforementioned in the previous section, our implementation is demonstrated based on ResNet, due to the fact that we observed the smallest FIDs when using ResNet as the backbone network. In order to evaluate the quality of the generated samples, we will show the results in qualitative and quantitive ways, i.e., sample visualization and FID. Sample visualization can reveal the detailed difference between the generated samples and the real samples. FIDs are known to be an effective measurement to judge sample quality. Under these two evaluations, the performance of WaveletGAN and its improvements to sample generation can be established.  

\subsection{Datasets}

The datasets utilized in the experiments include 
Fashion-MNIST \cite{Xiao2017Fashion} (a.k.a. FMNIST), 
KMNIST \cite{Clanuwat2018Kmnist}, 
and Street View House Numbers (SVHN) \cite{Netzer2011Reading}, which are all popular for benchmarking GANs. The configurations of all datasets are depicted in Tab. \ref{table:datasets}. 

FMNIST and KMNIST, which are both MNIST-like datasets, contain an equal number and scale of gray images in $28 \times 28$ pixels. The fashion styles and the Japanese letters all belong to 10 classes. The training and testing sets are divided in advanced, including 60,000 training samples and 10,000 test samples. Correspondingly, SVHN is an RGB image dataset, in which there are 73,257 training samples and 26,032 test samples in $32\times32$ pixels and with $3$ channels. Due to the fact that the input images are of different sizes, the GAN architectures are under distinct designs, as previously described in Sec. \ref{Method}. 

No matter using which datasets, the GANs  will generate fake images for the test samples after being trained using the training samples. The visualization, as well as the calculation of FIDs, will be conducted based on these generated images.  

\begin{table*}
  \caption{Datasets and their configurations.}
  \label{table:datasets}
  \center
  \begin{tabular}{llcrrc}
    \toprule
    Name			& Size 			& Classes & Train Set 	& Test Set 	& Samples  \\
    \midrule
    FMNIST \cite{Xiao2017Fashion}	& $28\times28$   	& 10 	& 60,000		& 10,000		
    & \raisebox{-0.7ex}{
       \includegraphics[width=0.6cm]{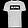}\ \includegraphics[width=0.6cm]{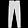} 
     \ \includegraphics[width=0.6cm]{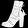}\ \includegraphics[width=0.6cm]{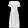}  }	\\
    KMNIST \cite{Clanuwat2018Kmnist}	& $28\times28$   	& 10 	& 60,000		& 10,000		
    & \raisebox{-0.7ex}{
       \includegraphics[width=0.6cm]{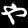}\ \includegraphics[width=0.6cm]{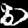}
     \ \includegraphics[width=0.6cm]{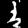}\ \includegraphics[width=0.6cm]{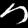} }	\\
    SVHN \cite{Netzer2011Reading}		& $32\times32\times3$   	& 10 	& 73,257		& 26,032		
    & \raisebox{-0.7ex}{
       \includegraphics[width=0.6cm]{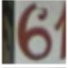}\ \includegraphics[width=0.6cm]{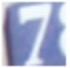} 
     \ \includegraphics[width=0.6cm]{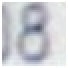}\ \includegraphics[width=0.6cm]{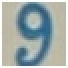} } 	\\
   \bottomrule
\end{tabular}
\end{table*}

\subsection{Generated samples}
\label{Generated}

\begin{figure*}
  \begin{center}
    \includegraphics[width=0.99\textwidth]{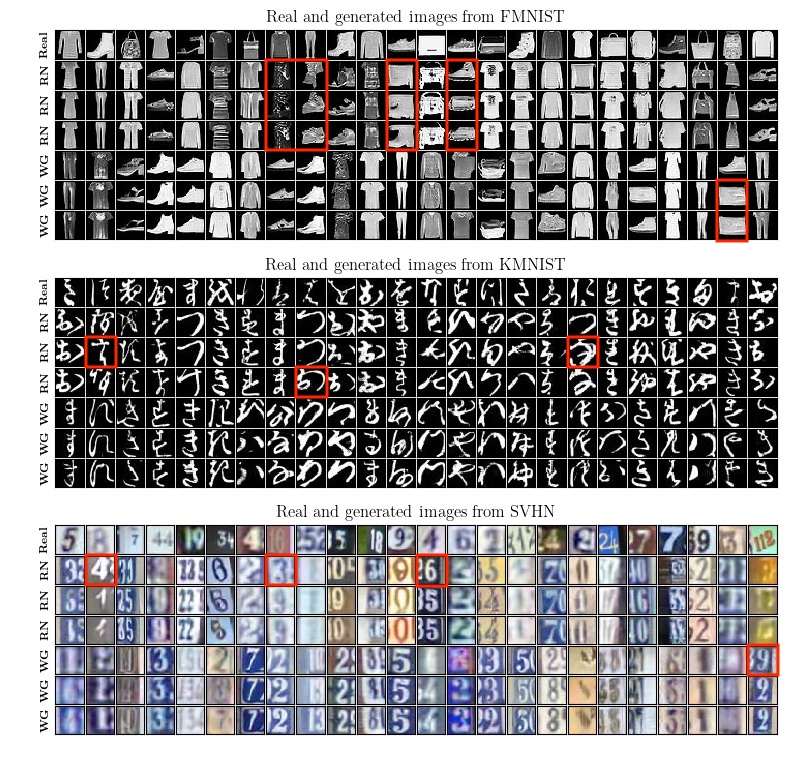}
  \end{center}
  \caption{\small  The real and generated samples from each dataset.}
  \label{fig:Genterated}
\end{figure*}

In order to intuitively evaluate WaveletGAN, we plot the generated samples in Fig. \ref{fig:Genterated}. Particularly, the first line shows some real samples (marked as Real) from each dataset for comparison. The following includes 3 lines of the fake samples (marked as RN) generated by the ResNet-based GAN and 3 lines of the generated samples (marked as WG) by our WaveletGAN. In this way, there are 7 lines of samples plotted for each dataset. It is noted that all the samples are directly picked up during the runtime of evaluating the models, which had no fixed order. The random appearance has less influence to judge their visualization in our opinion. For this reason, we do not reorder them manually.    

However, we manually highlight some of the visibly occluded samples with the red boxes. It is obviously that more samples generated by ResNet-based GAN (in line 2 to 4) suffer the problem than the ones outputted by our WaveletGAN. For example, many fake samples of shoe (in column 8, 9, 12, and 14 of line 2 to 4) from FMNIST are over occluded, or failed in image generation.  The same issue has been found in KMNIST and SVHN as well. That being said, WaveletGAN generates much fewer samples with these problems. This is merely a qualitative judgement, but we can continue to evaluate it through a quantitative metric.      

\subsection{Fréchet Inception Distances (FIDs)}
\label{FID}

Fréchet Inception Distance (FID) \cite{Heusel2017Gans} is one of the most significant quantitative metrics to evaluate the performance of GANs. Despite the fact that other measurements, e.g., Inception Score (IS) \cite{Salimans2016Improved} and GILBO \cite{Alemi2018Gilbo}, can also be utilized for this quantitative evaluation, FID is known to be popular and effective for the same job \cite{Chen2019Self}. Therefore, we consider FID as the mere metric in this work. Since our WaveletGAN is merely related with the outputted noise, it supports both unsupervised and conditional settings, technically, in most of the GAN architectures. In order to calculate FIDs, we generated the same number of fake samples, as the test samples, 3 times on each dataset, before computing the average FID for each case.

\begin{table}
  \caption{Best FIDs obtained in unsupervised (Un-cond.) and conditional  (Cond.) GANs.}
  \label{table:FIDs}
  \center
  \begin{tabular}{lrrrrrr}
    \noalign{\smallskip}
    & \multicolumn{2}{c}{FMNIST} & \multicolumn{2}{c}{KMNIST} & \multicolumn{2}{c}{SVHN} \\
    \toprule
    GANs 		& Un-cond. 	& Cond. & Un-cond. 	& Cond. & Un-cond. 	& Cond.  \\
    \midrule
    DC-GAN \cite{Radford2016DCGAN}			
    & 10.594  	&  10.149  	 	
    & 14.788 	&  12.573  	 	
    & 36.880  	&  41.172  	 \\
    WGAN-GP \cite{Gulrajani2017Improved} 		
    & 10.448  	&  10.680 	
    & 14.113  	&  13.169  	
    & 33.368  	&  33.500 	 \\
    SN-DCGAN \cite{Miyato2018Spectral} 			
    & 15.085	& 14.274 	
    & 11.200   	& 10.490  
    & 17.702  	& 17.402  \\
    SSGAN-ResNet \cite{Chen2019Self}
    & 23.165	& 20.062 	 	
    & 10.216   	& 10.918   	
    & 17.856 	& 17.733  	\\
    SSGAN-SNDCGAN \cite{Chen2019Self} 	
    & 13.895 	& 14.572  	
    & 14.084 	& 15.320 	
    & 17.004 	& 22.336 	 \\
    SN-ResNet \cite{Miyato2018Spectral}		
    & 8.320  	& 8.252  	
    & 4.983 	& 5.356  	
    & 16.267 	& 16.654  	 \\
    \midrule
    Our WaveletGAN 
    & 8.089			& \textbf{7.945}				
    & \textbf{4.874}	& 5.227				
    & \textbf{16.106} 	& 16.256  \\
    \noalign{\smallskip}
    \textit{~~Improvement Rate (\%)}  
    & $\uparrow$ 2.78 	& $\uparrow$ 3.73
    & $\uparrow$ 2.20 	& $\uparrow$ 2.42
    & $\uparrow$ 0.99 	& $\uparrow$ 2.39  \\
    \bottomrule 
    \noalign{\smallskip}
    \multicolumn{7}{l}{\textit{\textbf{Bold} values are the best (lowest) FIDs on the specific dataset.}} \\
    \multicolumn{7}{l}{\textit{Improvement Rate (\%) is calculated between WaveletGAN and SN-ResNet, e.g., on FMNIST, }} \\
    \multicolumn{7}{l}{\textit{~~the rate is $(8.320-8.089)/8.089=2.78\%$.}} \\
\end{tabular}
\end{table}


During the benchmarking, we compared our WaveletGAN with most of the state-of-the-art GANs, including  DC-GAN \cite{Radford2016DCGAN}, WGAN-GP \cite{Gulrajani2017Improved}, SN-DCGAN \cite{Miyato2018Spectral}, and SSGAN \cite{Chen2019Self}, under the unsupervised (without labels) and conditional (using labels \cite{Miyato2018Cgans}) settings.  Since our benchmarking experiments were conducted on \textit{compare\_gan}, we reused the same configurations from the tool, and set the parameters 
according to the GAN benchmark settings in \cite{Chen2019Self}. For example, the batch size is set to $64$ for all models, and the dimension of randomly initialized noise $\boldsymbol{z}_0$ is 128. The tool provides different structures of DC-GAN and SN-DCGAN, but spectral normalization (SN) \cite{Miyato2018Spectral} is designed as a configurable setting for all models. Therefore, we enabled SN on all models, including DC-GAN and WGAN-GP. For consistency, we set 5 iterations of sub-steps during each discriminator training step, and the hinge loss function was used in all GANs except SN-ResNet. Besides this, Adam was set as the optimizer for all models with a learning rate $\gamma=0.0002$. All models were trained in 100,000 steps. More detailed settings of each models can also be found in our accompanying code. 

The lowest FIDs by these selected GANs are recorded in Table \ref{table:FIDs}. Obviously, the smallest FIDs on all datasets are obtained by our WaveletGAN, as aforementioned, which utilizes ResNet as the backbone network. For example, the best FIDs are 4.874 (in \textbf{bold} style) obtained on KMNIST in unsupervised mode. The lowest FIDs on FMNIST and SVHN are 7.945 and 16.106 (in \textbf{bold} style), respectively. All three results are the lowest FIDs on the specific dataset. More impressively, these FIDs are much smaller than many other GANs, like DC-GAN, WGAN-GP and SN-DCGAN, which can only produce FIDs larger than 10. Furthermore, it can be observed that the recently proposed self-supervised GAN (SSGAN) generates FIDs 10.216 and over (on KMNIST), which are inferior to the results by our WaveletGAN. 

It is noted that whether using the labels or not this has a less decisive impact on the results. Conditional GANs produce lower FIDs most of the time, as shown in Table \ref{table:FIDs}. However, there are exceptions as well. For example, the unsupervised SN-ResNet outputs lower FIDs on KMNIST (4.983) and SVHN (16.267). This leads to the same results in our WaveletGAN, which obtained FIDs lower than the ones in the conditional setting. We focus on improving sample generation in GANs, rather than the behavior of using labels or not. The results demonstrate that wavelet-based filtering has nothing to do with this. The point is that WaveletGAN is good at generating high-fidelity samples according to the quantitive metric FID. 

\subsection{Improvements}
\label{Improvements}

Technically, the wavelet-based noise homogenization indeed supports any structures of GANs, but our implementations use ResNet as the backbone network (see Table \ref{table:resnetGAN}). This is because we observed the lowest FIDs in the SN-ResNet-based GAN implementation \cite{Miyato2018Spectral}. Therefore, we calculated the improvement rates between our WaveletGAN and SN-ResNet, as shown in the last line of Table \ref{table:FIDs}. 

Obviously, the highest improvements were obtained in conditional settings, where the most promising one is 3.73\% on FMNIST. The improvements on the other two datasets are greater than 2.39\%, which are equally impressive. Although the rate is slightly less than 1\% on SVHN in the unsupervised setting, 2.78\% and 2.20\% are observed on FMNIST and KMNIST, respectively. This demonstrates that noise homogenization by wavelet-based filtering consistently improves the sample generation in GANs.  

\section{Conclusions and Future Work}
\label{Conclusions}

The work proposes a novel GAN architecture that incorporates multi-channel wavelet filtering in the generator for the first time. Implemented as the \textit{WaveletDeconv} layer after the noise generation, which supports most of the GAN architectures, noise homogenization is activated and therefore enables high-fidelity sample generation in GANs. The implementation of WaveletGAN, based on the ResNet backbone, generates very high-fidelity fake samples, as well as the lowest FIDs, on multiple image datasets. Benchmarks through the open source GAN tool confirms that WaveletGAN has a superior performance than many state-of-the-art GANs.   

Despite this, there are still some known issues to be resolved. For example, wavelet transformation is a computationally-intensive operation, which calls for special accelerating techniques on GPU devices. The focus is about the generator, but it would be interested if the same mechanism can help to improve the discriminator. We will follow this idea and explore it as part of the future work.

\begin{ack}
We implemented our method based on the GAN benchmark tool provided by Google Research in https://github.com/google/compare\_gan. We want to thank the team for providing the initial code base. The authors acknowledge the support of the University of Macau (File no. MYRG2018-00053-FST), and NVIDIA Corporation with the donation of the Titan Xp GPU used for this research.
\end{ack}

\bibliographystyle{unsrt}
\bibliography{wavelet_gan}
\end{document}